\documentclass{article}

\usepackage[final]{neurips_2019}

\usepackage[utf8]{inputenc} 
\usepackage[T1]{fontenc}    
\usepackage{hyperref}       
\usepackage{url}            
\usepackage{booktabs}       
\usepackage{amsfonts}       
\usepackage{nicefrac}       
\usepackage{microtype}      
\usepackage{comment}
\usepackage{ragged2e}
\usepackage{fixltx2e}
\usepackage{caption}
\usepackage{subcaption}
\usepackage{textgreek}
\usepackage[]{natbib}
\title{NLP meets psychotherapy: Using predicted client emotions and self-reported client emotions to measure emotional coherence}

\author{%
    Neha Warikoo $^{1}$ , Tobias Mayer $^{1}$, Dana Atzil-Slonim $^{2}$, Amir Eliassaf $^{2}$,  Shira Haimovitz $^{2}$, Iryna Gurevych $^{1}$ \\ \\
   $^1$Ubiquitous Knowledge Processing Lab (UKP Lab) \\
   Department of Computer Science and Hessian Center for AI (hessian.AI) \\
   Technical University of Darmstadt \\
   \url{www.ukp.tu-darmstadt.de} \\ \\
   $^2$ Psychotherapy Research Lab (PR Lab) \\
   Department of Psychology\\
   Bar-Ilan University\\
   \url{www.prlab.co.il/en/} \\ \\
}

\begin{document}

\maketitle

\begin{abstract}
Emotions are experienced and expressed through various response systems. Coherence between emotional experience and emotional expression is considered highly important to clients' well being. To date, emotional coherence has been studied at a single time point using lab-based tasks with relatively small datasets. No study has examined emotional coherence between the subjective experience of emotions and utterance-level emotions over therapy sessions or whether this coherence is associated with clients' well being. Natural language Processing (NLP) approaches have been applied to identify emotions during psychotherapy dialogue, which can be implemented to study emotional processes on a larger scale and with specificity. However, these methods have yet to be used to study coherence between emotional experience and emotional expression over the course of therapy and whether it relates to clients' well-being.\\
This work presents an end-to-end approach where we use emotion predictions from our transformer based emotion recognition model to study emotional coherence and its diagnostic potential in psychotherapy research. We first employ our transformer based approach on a Hebrew psychotherapy dataset to automatically label clients' emotions at the utterance level in psychotherapy dialogues. We subsequently investigate the emotional coherence between clients' self-reported emotional states and our model-based emotion predictions. We also examine the association between emotional coherence and clients' well being.\\
The findings indicate a significant correlation between clients' self-reported emotions and positive and negative emotions expressed verbally during psychotherapy sessions. Coherence in positive emotions was also highly correlated with clients well-being. These results illustrate how NLP can be applied to identify important emotional processes in psychotherapy to improve diagnosis and treatment for clients who suffer from mental-health problems.  
\end{abstract}

\section{Introduction}
One of the main goals of psychotherapy is to help clients develop greater coherence between their emotional experiences and the words that describe them, since this can help them respond more effectively to emotional events \citep{fosha2001dyadic, greenberg2012emotions}. Emotional coherence, which is defined as the coordination between different emotional responses as the emotion unfolds over time, is considered crucial to individuals' well-being \citep{levenson2014autonomic, mauss2005tie}. \\
Most previous studies assessing emotional coherence have used a lab-based emotional stimuli to evaluate emotional responses (e.g. \citep{hastings2009dysregulated, negrao2005shame, wagner2003emotional}) 
, which cannot capture a genuine real-life emotional response to an interpersonal interaction (such as between clients and therapists). In addition, most of these studies have compared clinical to non-clinical population at a single time point, which limits their ability to determine how different emotional responses change together over time across psychotherapy sessions. Furthermore, these studies focus on several emotional responses, such as between individuals’ subjective reports and physiological response \citep{brown2020coherence, lohani2018emotional} or facial expressions \citep{lohani2018emotional}. \\ 
Psychotherapy researchers have highlighted the importance of coherence between subjective emotional experiences and the words that describe them, since it allows people to give meaning to their experiences which may lead to a better ability to communicate and regulate emotions in a more adaptive way \citep{greenberg2012emotions, lane2015memory}. However, to the best of our knowledge, no empirical study has tested the coherence between these two emotional responses. \\
Lack of research in this direction may be  due to the fact that until recently psychotherapy studies were dependent on human annotators, which limited their ability to capture within-session emotional processes on a large scale (e.g., \citep{imel2015computational}). Studies in the last few year, however, have demonstrated the value of implementing Natural Language Processing (NLP) to understand psychotherapy dialogue \citep{Tanana2021HowDY, gibson2017attention}. Given the vast potential of NLP in data-driven analysis it has also become a key resource for psychotherapy diagnosis as reviewed by \citep{shatte2019machine}.\\
NLP methods that can automatically capture utterance-level emotions at a higher scale and specificity could successfully address the current gaps in the emotional coherence literature and examine emotional coherence in channels that have yet to be investigated. Exploring whether emotional coherence is associated with clients' well-being over the course of therapy may also lead to important conclusions about the diagnostic value of emotional coherence. This work describes an end-to-end approach where a transformer-based Hebrew language model was used to develop utterance-level predictions of client emotions. These automated emotion predictions were then used to study emotional coherence and its associations with client well-being measures. The main contributions of this work are:
\begin{itemize}
    \item This is the first study to assess the emotional coherence between the subjective experience of client emotions and the verbal expression of their emotions.
    \item The automated predictions from the transformer model can be applied to scale-up psychotherapy research and investigate therapy dialogue at the utterance-level.
    \item The association between the clients' subjective experience of emotions and the model's predictions of clients' emotions validates the ability of this approach to effectively capture clients' emotional processes in therapy dialogues.
    \item Compared to previous studies that have assessed emotional coherence at one time point and in lab-based tasks, the current study used longitudinal data to assess how emotional coherence unfolds from session to session in genuine interactions between clients and therapists.
\end{itemize}

\section{Methods}
The following hypotheses were examined: 
\begin{enumerate}
    \item There should be a positive correlation between verbal expression of emotions and  clients' subjective experience of emotions. 
    \item Higher emotional coherence should be associated with clients' better well-being.
\end{enumerate}

\subsection{Dataset}
\label{section:dataset}
We study a data-set comprised of 872 sessions collected as part of a university clinic’s regular practice of monitoring clients’ progress \citep{atzil2021using}. The language of therapy was Modern Hebrew. A total of 196 sessions from the original dataset were annotated for emotion labels at the utterance level by clinical experts i.e. \textbf{BIU-872\_Gold}. These emotion labels were coded as: \textit{Negative}, \textit{Positive}, \textit{Neutral:} which was defined as neither positive nor negative emotions and \textit{Mixed:} which was defined as both positive and negative emotions \citep{greenberg2012emotions}. The remainder of the un-annotated sessions were called \textbf{BIU-872\_Silver}. \\
After each session, clients self-reported their emotional experience using the Profile of Mood States (POMS; \citep{cranford2006procedure}). The POMS consists of 12 words aggregated to describe current negative (e.g, sad) or positive (e.g., happy) emotional states.  At the beginning of each session, the clients also completed the Outcome Rating Scale (ORS; \citep{miller2003outcome}) to assess their well-being. It consists of four visual analog scale ranging from 0 to 10.

\subsection{Labelling Client Emotions for BIU-872\_Silver}

\subsubsection{Task Definition}
\label{section:task_definition}
BIU-872 is a conversation dataset with two speakers i.e. \textit{the client} (C) and \textit{the therapist} (T) and the task is to label client emotions at the utterance-level. Formally, given an input sequence of \textit{N} utterances [u\textsubscript{1}\textsuperscript{p}, u\textsubscript{2}\textsuperscript{p}.......u\textsubscript{N}\textsuperscript{p}], where p=[C, T] and each utterance u\textsubscript{i}\textsuperscript{p}=[u\textsubscript{i,1}, u\textsubscript{i,2},........u\textsubscript{i,T}] has \textit{T} words \textit{u\textsubscript{i,j}} spoken by party \textit{p}, the task is to label emotions only for client utterances [u\textsubscript{1}\textsuperscript{C}, u\textsubscript{2}\textsuperscript{C}.......u\textsubscript{N}\textsuperscript{C}].
\subsubsection{Model}
\label{section:model}
To study client emotions, we adapted AlephBERT on a downstream classification task of emotion recognition (ER). We chose AlephBERT for domain training because a) it achieves state of the art (SOTA) results on benchmark datasets for HebrewNLP tasks b) it is pre-trained on a large Hebrew corpus of 52K \citep{seker2021alephbert} and we are also working with a dialogue dataset in Modern Hebrew. \\
We trained AlephBERT on BIU-872\_Gold over 10-fold cross validation to account for the variability in results. Hyper-parameter tuning was done on development dataset (10\% of BIU-872\_Gold per fold). The learning rate was set to 2e-6 with L2 regularization (decay=0.0001). The maximum token size \textit{T} per utterance was set to 128 and partial class balance was implemented due to skewed emotion label distribution \citep{chawla2002smote}. In line with previous ER studies in psychotherapy, we evaluated the results from the trained model with the F1 micro \citep{Tanana2021HowDY}. We then used our BIU-872\_Gold trained AlephBERT model to label client emotions for BIU-872\_Silver.

\subsection{Emotional Coherence as a diagnostic tool}

\subsubsection{POMS vs Utterance-level emotions}
\label{section:coherence_measure}
In the next step, we tested \textit{hypothesis 1} which evaluated coherence between the self-reported client emotions i.e. POMS \textbf{P\textsubscript{e}\textsuperscript{m}} (cumulative) and utterance-level (i.e. verbal expression) client emotions \textbf{U\textsubscript{e}\textsuperscript{m}} (normalized), where e = emotion labels (pos, neg) and m = session number. As mentioned in Section \ref{section:dataset}, there are four utterance level coding in the dataset, but we only focus on \textit{positive} and \textit{negative} emotions to study any meaningful change in patient behavior.\\
Total utterance size and emotion labels varied in count both within and across sessions. To reduce the disparity between long and short sessions and to understand the real significance of each emotion throughout session \textit{m}, we performed emotion label normalization for each session as follows:
\begin{equation}
    U\textsubscript{e}\textsuperscript{m} = \frac{\#u\textsubscript{e}\textsuperscript{m}}
    {\sum_{i\subset[pos,neg,mix, neu]}\#u\textsubscript{i}\textsuperscript{m}}
\end{equation}
where \#u\textsubscript{e}\textsuperscript{m} is the number of emotion \textit{e} across all utterances for session \textit{m}.\\
The client POMS was evaluated on six different emotion sub-scales. We collated these sub-scales for \textit{pos}=[calmness,contentment,vigor] and \textit{neg}=[anger,sad,anxiety] emotions to study POMS in a binary emotion setting:
\begin{equation}
    P\textsubscript{e}\textsuperscript{m} = \sum_{e\subset[pos, neg]}p\textsubscript{e}\textsuperscript{m}
\end{equation}
where p\textsubscript{e}\textsuperscript{m} is the total score for emotion sub-scale \textit{e} for the entire session \textit{m}.\\
Coherence between the POMS and utterance-level emotions was measured for both positive and negative emotions. We used the Pearson implementation from the Scipy library to calculate the correlations i.e. Cohr(U\textsubscript{e}, P \textsubscript{e}) where \textit{e}=[pos, neg] and the corresponding significance values \citep{benesty2009pearson, kowalski1972effects, virtanen2020scipy}. Coherence experiments with BIU-872\_Gold aimed at testing of \textit{hypothesis 1} by using expert coded emotion labels as the verbal emotion expression. Then we examined emotional coherence on BIU-872\_Silver for application results, where the emotion labels are developed from the transformer-based model. Significance cutoff for correlation results was set at \textalpha =0.05.

\subsubsection{Association between Emotional Coherence and ORS}
\label{section:coherence_ORS}
In order to test \textit{hypothesis 2}, we calculated correlation between emotional coherence and ORS at the client-level across treatment. The session-wide results for each client \textit{'l'} were first summarized using mean ({\textmu}) and then the Pearson correlation was used to calculate the association i.e. Corr({\textmu}(Cohr(U\textsubscript{e}\textsuperscript{l}, P \textsubscript{e}\textsuperscript{l})), {\textmu}(ORS\textsubscript{e}\textsuperscript{l})), where e=[pos, neg] and l=client\_id.

\section{Results and Discussion}
\subsubsection{Emotion labels for BIU-872\_Silver}
The test data for BIU-872\_Gold prediction results from the transformer model achieved a moderate F1 of 0.66 for ER in psychotherapy. We adapted the pre-trained model to develop utterance-level emotion labels for BIU-872\_Silver. The pre-trained ER model allows psychotherapy dialogues to be studied with more granularity at the dialogue-level and on a larger scale than previously examined. While in the current study this model was used to explore emotional coherence, the automatic labeling of emotions also opens up the possibilities to study a wide range of other emotional processes in psychotherapy.

\begin{table}
\begin{subtable}[c]{0.6\textwidth}
\centering
\begin{tabular}{c|c|c} 
    & BIU-872\_Gold & BIU-872\_Silver \\ \hline\hline
    (P\textsubscript{pos}, U\textsubscript{pos}) & (0.29,7.8e-05) & (\textbf{0.27, 4.3e-12}) \\ 
    (P\textsubscript{neg}, U\textsubscript{neg}) & (0.24, 0.001) & (\textbf{0.21, 4.1e-08}) \\ \hline
    \end{tabular}
\subcaption{Session-wide Correlation between POMS and Utterance \\ emotion labels}
\label{tab:emotional_coherence}
\end{subtable}
\begin{subtable}[c]{0.4\textwidth}
\centering
\begin{tabular}{c|c} 
    & ORS \\ \hline \hline
    Cohr(U\textsubscript{pos}, P\textsubscript{pos}) & (\textbf{0.67, 0.048}) \\ 
    Cohr(U\textsubscript{neg}, P\textsubscript{neg})) & (-0.37, 0.321)  \\ \hline
    \end{tabular}
\subcaption{Client-level correlation between \\ Emotional coherence and ORS}
\label{tab:coherenceORS}
\end{subtable}
\caption{Correlation analysis on BIU-872}
\end{table}

\subsubsection{Emotional Coherence between self-reports and verbal expression}
The results on BIU-872\_Gold in Table \ref{tab:emotional_coherence} show there is a significant and positive correlation between P\textsubscript{pos} and U\textsubscript{pos} (0.29) and P\textsubscript{neg} and U\textsubscript{neg} (0.24) across all the sessions. This result was based on expert annotated emotion labels, and therefore it validates our \textit{hypothesis 1} using traditional psychotherapy analysis. These results are consistent with previous studies that have reported coherence across various emotional response systems (e.g., \citep{brown2020coherence}), but extend beyond them by showing that coherence also occurs between emotional experience and verbal emotion expression. \\
Table \ref{tab:emotional_coherence} also depicts a significant positive correlation between P\textsubscript{pos} and U\textsubscript{pos} (0.27) and P\textsubscript{neg} and U\textsubscript{neg} (0.21) for BIU-872\_Silver. These results validate the performance of the transformer-based ER approach to automatically detect genuine emotions from text with specificity. This result further underscores the usefulness of this transformer-based model in detecting emotional coherence on a larger scale. 

\subsubsection{Association between emotional coherence and well being}
The results in Table \ref{tab:coherenceORS} indicate significant results (0.048) with a high correlation (0.67) between Cohr(U\textsubscript{pos}, P\textsubscript{pos}) and the ORS measure. This finding supports a largely untested theoretical claim that the emotional coherence between subjective experience and the verbal expression of emotions is important for clients' well-being \citep{greenberg2012emotions, lane2015memory}. The findings of association between emotional coherence and well-being only for positive emotions is in line with previous studies reporting similar results with other emotional channels \citep{mauss2005tie}. \\ 
One possible explanation for this finding is that negative emotions tend to be more salient than positive emotions \citep{baumeister2001bad}, and hence may be more easily recognized. However, better well-being appears to be achieved when clients express and at the same time recognize their positive emotions. \\
Our results highlight that by using NLP-based ER models, clinicians may be better able to identify clients or sessions characterized by a high dissociation between emotional experience and verbal expression of emotions and direct their interventions to help clients accordingly.

\section*{Ethics Statement}
The materials were only collected after securing approval from the authors’ university ethics committee. Only clients who gave their consent to participate were included in the study. Clients were told that they could choose to terminate their participation in the study at any time without jeopardizing treatment. All sessions were audiotaped and transcribed according to a protocol ensuring confidentiality and masking of any identifying information, such as names and places. Finally, to ensure privacy due to the sensitive nature of our data, secured servers were used with limited access to develop this study.

\subsubsection*{Acknowledgments}
From UKP Lab, this work has been funded by the LOEWE Distinguished Chair “Ubiquitous Knowledge Processing” (LOEWE initiative, Hesse, Germany). From PR Lab, this work was supported by the Israel Science Foundation, (ISF \#2466/21 to Dana Atzil-Slonim).

\bibliography{neurips_2019}
\small

\end{document}